%% file: main.tex
\documentclass[runningheads]{llncs}

\usepackage[mobile]{eccv}

\usepackage{eccvabbrv}

\usepackage{graphicx}
\usepackage{booktabs}
\usepackage{xcolor}
\usepackage{multirow}
\usepackage{amssymb}    %
\usepackage{wrapfig}
\usepackage{marvosym}
\usepackage{url}

\definecolor{nbarrier}{RGB}{255, 120, 50}
\definecolor{nbicycle}{RGB}{255, 192, 203}
\definecolor{nbus}{RGB}{255, 255, 0}
\definecolor{ncar}{RGB}{0, 150, 245}
\definecolor{nconstruct}{RGB}{0, 255, 255}
\definecolor{nmotor}{RGB}{200, 180, 0}
\definecolor{npedestrian}{RGB}{255, 0, 0}
\definecolor{ntraffic}{RGB}{255, 240, 150}
\definecolor{ntrailer}{RGB}{135, 60, 0}
\definecolor{ntruck}{RGB}{160, 32, 240}
\definecolor{ndriveable}{RGB}{255, 0, 255}
\definecolor{nother}{RGB}{139, 137, 137}
\definecolor{nsidewalk}{RGB}{75, 0, 75}
\definecolor{nterrain}{RGB}{150, 240, 80}
\definecolor{nmanmade}{RGB}{213, 213, 213}
\definecolor{nvegetation}{RGB}{0, 175, 0}

\definecolor{nvcolor}{RGB}{119,185,0}
\definecolor{roadcolor}{RGB}{234,51,246}
\definecolor{sidewalkcolor}{RGB}{68,8,72}
\definecolor{parkingcolor}{RGB}{241,156,249}
\definecolor{othergroundcolor}{RGB}{160,32,76}
\definecolor{buildingcolor}{RGB}{246,202,69}
\definecolor{carcolor}{RGB}{111,149,238}
\definecolor{truckcolor}{RGB}{74,32,172}
\definecolor{bicyclecolor}{RGB}{136,227,242}
\definecolor{motorcyclecolor}{RGB}{37,59,146}
\definecolor{othervehiclecolor}{RGB}{96,81,242}
\definecolor{vegetationcolor}{RGB}{79, 173, 50}
\definecolor{trunkcolor}{RGB}{126, 65, 22}
\definecolor{terraincolor}{RGB}{171, 238, 105}
\definecolor{personcolor}{RGB}{234, 60, 49}
\definecolor{bicyclistcolor}{RGB}{234, 66, 195}
\definecolor{motorcyclistcolor}{RGB}{138, 42, 90}
\definecolor{fencecolor}{RGB}{238, 128, 69}
\definecolor{polecolor}{RGB}{252, 241, 161}
\definecolor{trafficsigncolor}{RGB}{233, 51, 35}
\definecolor{other-struct.color}{RGB}{255, 150, 0}
\definecolor{other-objectcolor}{RGB}{50, 255, 255}
\definecolor{lane-markingcolor}{RGB}{150, 255, 170}
\definecolor{color1}{RGB}{176, 36, 24}
\definecolor{color2}{RGB}{0, 176, 80}
\definecolor{color3}{RGB}{0, 0, 200}

\usepackage[accsupp]{axessibility}  %

\usepackage[colorlinks=true,urlcolor=magenta]{hyperref}

\usepackage{ragged2e}

\usepackage{orcidlink}

\begin{document}

\title{DriveTok: 3D Driving Scene Tokenization for Unified Multi-View Reconstruction and Understanding} 

\titlerunning{DriveTok for Scene Understanding}

\author{
Dong Zhuo\inst{1}$^{,*}$ \and
Wenzhao Zheng\inst{1}$^{,*,\dagger}$ \and
Sicheng Zuo\inst{1}  \\
Siming Yan\inst{2} \and Lu Hou\inst{2}
\and Jie Zhou\inst{1} \and Jiwen Lu\inst{1}
}

\begingroup
\renewcommand\thefootnote{}\footnotetext{$^{*}$ Equal contributions. $^{\dagger}$ Project leader.}
\endgroup

\authorrunning{D. Zhuo et al.}

\institute{
\textsuperscript{1} Tsinghua University \quad
\textsuperscript{2} Yinwang Intelligent Technology Co. Ltd.\\
Page: \url{https://paryi555.github.io/DriveTok/} \\ 
Code: \url{https://github.com/paryi555/DriveTok}
}

\renewcommand\twocolumn[1][]{#1}%
\maketitle

\vspace{-5mm}
\begin{center}
     \centering
     \includegraphics[width=1\linewidth]{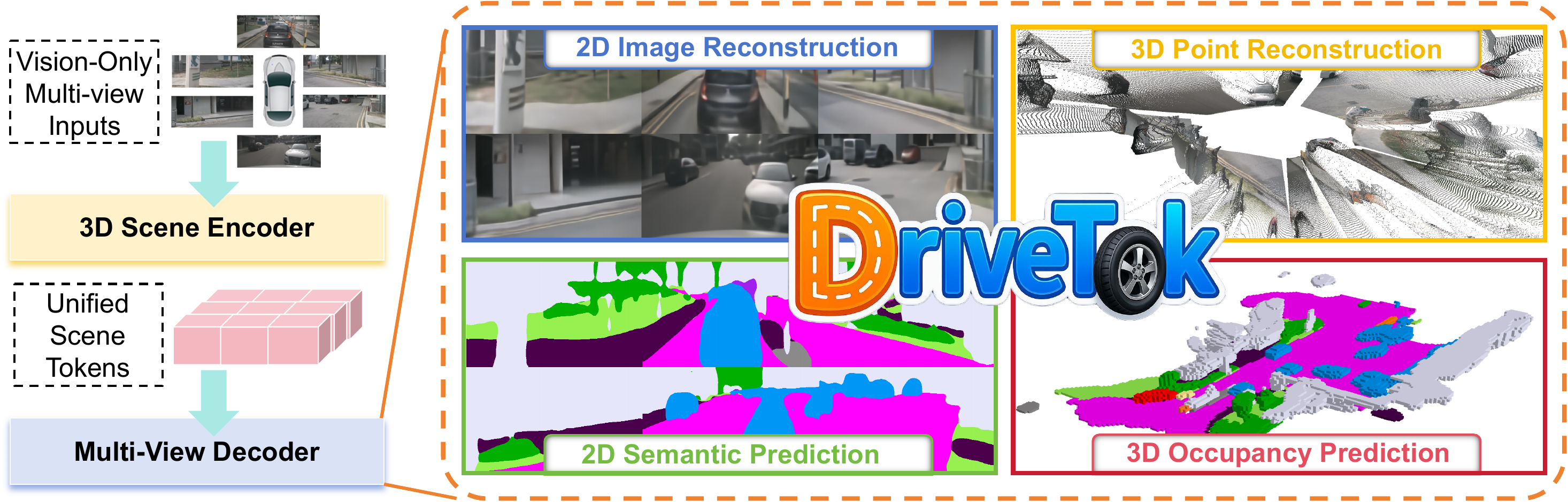}
     \vspace{-6mm}
     \captionof{figure}{    
     We propose our DriveTok for multi-view scene reconstruction and understanding. Multi-view images are processed by a 3D scene encoder to produce unified scene tokens, independent of camera layout and resolution. A spatial-aware multi-view decoder renders predictions in both image and occ spaces. Through joint multi-task training, our scene tokens encode rich textural, semantic, and geometric information. 
     }
\label{fig:teaser}
\vspace{-3mm}
\end{center}%

\input{sec/0_abstract}

\input{sec/1_introduction}

\input{sec/2_related}

\input{sec/3_method}

\input{sec/4_experiment}

\input{sec/5_conclusion}

\bibliographystyle{splncs04}
\bibliography{main.bbl}
\clearpage
\input{sec/X_suppl}
\end{document}

%% file: sec/0_abstract.tex
\begin{abstract}
With the growing adoption of vision-language-action models and world models in autonomous driving systems, scalable image tokenization becomes crucial as the interface for the visual modality.
However, most existing tokenizers are designed for monocular and 2D scenes, leading to inefficiency and inter-view inconsistency when applied to high-resolution multi-view driving scenes. 
To address this, we propose DriveTok, an efficient 3D driving scene tokenizer for unified multi-view reconstruction and understanding. 
DriveTok first obtains semantically rich visual features from vision foundation models and then transforms them into the scene tokens with 3D deformable cross-attention. 
For decoding, we employ a multi-view transformer to reconstruct multi-view features from the scene tokens and use multiple heads to obtain RGB, depth, and semantic reconstructions.
We also add a 3D head directly on the scene tokens for 3D semantic occupancy prediction for better spatial awareness.
With the multiple training objectives, DriveTok learns unified scene tokens that integrate semantic, geometric, and textural information for efficient multi-view tokenization.
Extensive experiments on the widely used nuScenes dataset demonstrate that the scene tokens from DriveTok perform well on image reconstruction, semantic segmentation, depth prediction, and 3D occupancy prediction tasks.
\end{abstract}

%% file: sec/1_introduction.tex
\section{Introduction}
\label{sec:intro}
The field of autonomous driving is undergoing a shift from a perception-centric pipeline toward a reasoning-based pipeline for complex driving scenes~\cite{mao2023gpt, shao2024lmdrive, xu2024drivegpt4}. 
This transition is largely driven by vision-language-action models (VLAs)~\cite{wang2002orion, zhou2025opendrivevla, jiang2025diffvla, zhou2025autovla, li2025drivevla} and world models~\cite{hu2023gaia, wang2024drivedreamer, zhao2025drivedreamer, wang2024driving, zheng2024doe, li2025drivevla},
which are increasingly applied to enhance the cognitive capabilities of autonomous systems. 

A central challenge in this evolution is how to represent the sensor inputs with comprehensive (i.e., low-level information for reconstruction) yet also semantically rich (i.e., high-level information for understanding) representations.
This information is critical to allowing large models to reason about possible events, interpret context, and respond robustly to unforeseen situations~\cite{sani2024graph}. 
This makes the design of a good visual scene tokenizer important for driving systems.
However, most existing visual tokenizers focus on monocular and generic-domain image reconstruction and usually tokenize inputs into per-image 2D patch tokens.
They cannot capture the spatial 3D information that is vital for driving.
Also, autonomous vehicles are typically equipped with surrounding high-resolution cameras, and per-image tokenization leads to a large set of tokens and thus inefficiency for the subsequent large models.

To address these challenges, we propose DriveTok, an efficient 3D scene tokenizer tailored for autonomous driving systems. DriveTok utilizes a unified scene representation to transform inputs from multiple high-resolution cameras into a fixed number of geometry-aware scene tokens. These scene tokens are agnostic to image resolution and camera count, ensuring consistent multi-view reasoning and efficient visual tokenization. Semantic-rich visual features are first extracted using a pre-trained vision foundation model, then mapped to 3D space through 3D deformable cross-attention. For decoding, a multi-view transformer reconstructs RGB, depth, and semantic maps, while an additional 3D head directly predicts 3D semantic occupancy from the scene tokens. These multi-task objectives enable DriveTok to learn unified scene tokens that encode semantic, geometric, and textural information. Extensive experiments on the nuScenes dataset~\cite{caesar2020nuscenes} validate the effectiveness of these tokens across image reconstruction, semantic segmentation, depth prediction, and 3D occupancy prediction tasks.

%% file: sec/2_related.tex
\section{Related Work}
\label{sec:related}
\textbf{Scene Representations for Autonomous Driving.}
Conventional methods employs depth maps to represent structural information for downstream driving tasks~\cite{zhou2017unsupervised,godard2019digging,chang2018pyramid,wang2019pseudo,ranftl2020towards,ranftl2021vision}.
They fail to capture an explicit 3D scene layout, limiting cross-view geometric consistency and global planning.
To obtain an explicit 3D scene representation from multi-camera images, some methods proposed to lift image features to the Bird’s-Eye-View (BEV)~\cite{philion2020lift, huang2021bevdet, li2023bevdepth}. 
For example, Lift-Splat-Shoot (LSS)~\cite{philion2020lift} back-projects per-pixel distributions to 3D then aggregates in BEV. 
BEVFormer~\cite{li2024bevformer} introduces spatio-temporal transformers with multi-scale deformable attention to sample informative regions efficiently. 
They improve 3D spatial awareness and downstream detection/segmentation, especially in surround-view setups. 
Beyond a single top-down plane, TPVFormer~\cite{huang2023tri} augments BEV with two orthogonal vertical planes (Tri-Perspective View) to better capture fine 3D structures that a single BEV plane struggles to encode.

Recent methods model scenes as sparse sets of parametric primitives that emphasize objects while avoiding dense volumetric grids. 
GaussianFormer~\cite{huang2024gaussianformer} proposed to represent scenes with sparse 3D semantic Gaussians to predict 3D occupancy more efficiently than voxel grids. 
QuadricFormer~\cite{zuo2025quadricformer} further replaces Gaussians with the more flexible superquadrics, leveraging richer shape priors to improve efficiency.
However, most existing 3D scene representations often focus on optimizing the performance on specific tasks, which deviates from the emerging demand to learn comprehensive (i.e., for reconstruction) yet semantically rich (i.e., for understanding) scene tokens.

\textbf{Visual Tokenizers.} Recent advancements in visual tokenization have primarily focused on compressing images into discrete or continuous latent tokens for downstream generation and reconstruction. Pioneering approaches like VQ-VAE~\cite{van2017neural} and and VQ-GAN~\cite{esser2021taming} introduce quantized latent spaces optimized for high-fidelity image generation. More recent works, including SEED~\cite{ge2307planting},  TiTok~\cite{yu2024image} and Janus~\cite{wu2025janus}, push toward autoregressive or unified multimodal frameworks, enabling tasks like image generation or instruction-following. Other innovations such as MetaMorph~\cite{tong2025metamorph}, VA-VAE~\cite{yao2025reconstruction}, and FlowMo~\cite{sargent2025flow} improve reconstruction and alignment by incorporating foundation models, diffusion priors, or optimization techniques. 

However, most existing tokenizers are trained on generic web-scale datasets and operate in a per-image fashion, without accounting for the spatial consistency and geometric structure crucial in autonomous driving. As a result, they struggle with domain shift and lack the ability to model cross-view 3D layouts. Although recent works such as BEV-VAE~\cite{chen2025bev} and triplane-based representations~\cite{ivanovic2025efficient} have explored geometry-aware latent spaces for driving scenes, they remain limited in semantic richness and scene-level understanding. BEV-VAE encodes low-resolution multi-view inputs into BEV-aligned latent features, capturing geometry but neglecting fine-grained semantics. The triplane-based tokenizer achieves efficient multi-camera representation through 3D volumetric encoding but focuses primarily on policy learning without explicit semantic modeling.
In contrast, our DriveTok introduces a unified scene tokenization framework designed for multi-view driving scenes. By transforming multi-view inputs into resolution- and camera-agnostic scene tokens using 3D deformable attention, DriveTok ensures strong spatial awareness. Its joint training over image-space and 3D-space objectives leads to semantic-rich scene tokens that better align with downstream vision-language models for holistic understanding and reasoning in autonomous driving scenes.

%% file: sec/3_method.tex
\begin{figure*}[t]
    \centering
    \includegraphics[width=\linewidth]{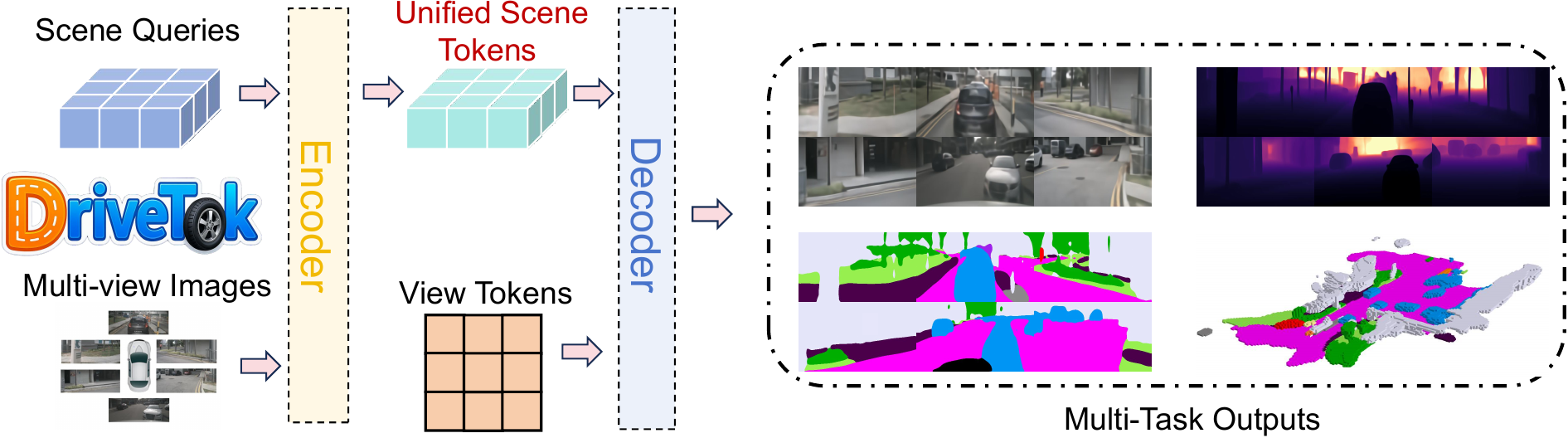}
    \vspace{-7mm}
    \caption{\textbf{Illustration of DriveTok.} DriveTok processes multi-view images and scene queries through an encoder-decoder architecture to produce unified scene tokens and generate diverse autonomous driving scene reconstruction and understanding outputs.}
    \label{fig:module}
    \vspace{-7mm}
\end{figure*}

\vspace{-3mm}
\section{Proposed Approach}
\label{sec:method}
\begin{figure*}[t]
     \centering
     \includegraphics[width=1\textwidth]{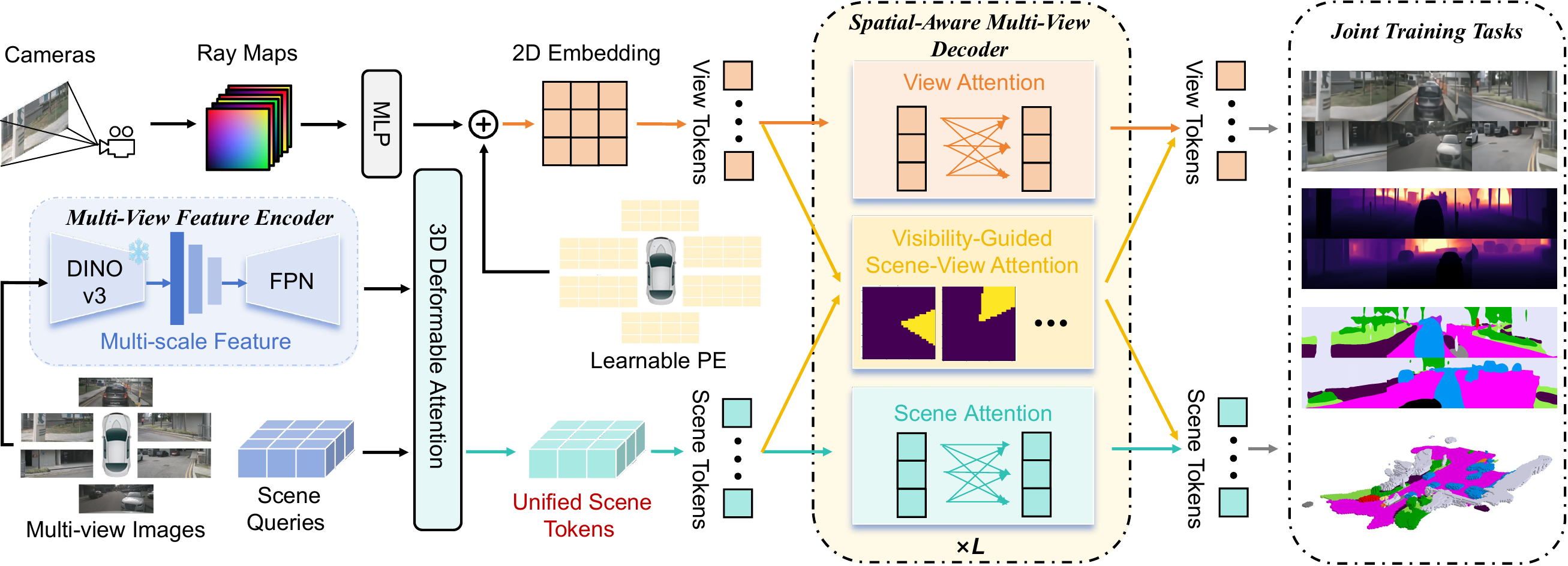}
     \vspace{-7mm}
     \captionof{figure}{\textbf{Overview of DriveTok.} Surround-view images are encoded by a 3D scene encoder. view tokens (with learnable and Plücker-ray embeddings) and scene tokens interact through a spatial-aware multi-view transformer with visibility-guided scene-view attention. Joint pretraining uses image reconstruction, depth prediction, semantic prediction, and occupancy prediction.
 }
 \label{fig:overview}
     \vspace{-5mm}
 \end{figure*}

The overall architecture of our method is illustrated in \cref{fig:overview}. 
Our framework consists of three main modules: a semantic-aware scene encoder, a spatial-aware multi-view transformer, and a set of task heads. 
This pipeline produces unified scene tokens that integrate appearance, semantics, and geometry from inputs, paving the way for unified autonomous driving.

\vspace{-1mm}
\label{method}
\subsection{3D Driving Scene Tokenization}
Conventional image tokenizers typically operate in the 2D image space, compressing individual images $I_i \in \mathbb{R}^{H \times W \times 3}$ into latent or discrete sequences $Z_i \in \mathbb{R}^{L \times d}$ optimized for image-level reconstruction or generation, Where $L$ is the length of sequence and $d$ is the embedding dimension. However, when applied to autonomous driving, this per-image formulation is fundamentally limited. Multi-camera inputs $\{I_i\}_{i=1}^N$ are processed independently, neglecting their shared 3D structure and resulting in view-inconsistent tokens $\{Z_i\}^{N}_{i=1}$ that lack spatial alignment and semantic grounding. Moreover, tokenizing each high-resolution image separately incurs computational cost proportional to $\mathcal{O}(N \cdot H \cdot W)$ and redundant representations in overlapping regions, making it inefficient and poorly scalable for dense, multi-view driving scenes.  

To address these limitations, we propose a 3D driving scene tokenization framework as shown in \cref{fig:module}, which lifts multi-view image features into a unified 3D space to generate compact scene tokens $B$. Image features extracted from a vision foundation model are projected into a global scene grid $\mathcal{Q}$ using geometry-aware sampling, where information from all camera views is aggregated. These scene tokens are spatially consistent, resolution-agnostic, and encode a combination of semantic, geometric, and visual cues, enhanced through our joint multi-task training strategy. Our approach ensures coherent cross-view understanding and produces a compact, structured representation that is decoupled from both input resolution and the number of cameras.

\vspace{-1mm}
\subsection{Vision Foundation Model as Scene Encoder}
As evidenced by \cite{zheng2025vision}, building an image tokenizer on top of a pretrained vision foundation model leads to stronger semantic representations. Such semantics are essential for autonomous driving systems to achieve robust scene understanding and to support safer, more human-aligned decision making. Accordingly, we adopt a pretrained vision foundation model as our image encoder to ensure strong generalization and semantic richness in the extracted features.

Specifically, given surround-view images $\{I_i\}^N_{i=1}$ at time $t$, we extract semantically and texturally rich per-camera features using a pretrained vision foundation backbone and an FPN~\cite{lin2017feature}, producing $F_i = \text{ImageEnc}(I_i) \in \mathbb{R}^{H_f \times W_f \times C}$, which serve as the basis for generating the unified scene representation. We then map these features into a fixed scene grid $\mathcal{Q} = \{ q_{ij} | i=1,...,W_b, j=1,...,H_b\}$ using a scene query lifter followed by \cite{li2024bevformer} with multi-scale deformable attention, which leverages the intrinsic/extrinsic of each camera to sample informative image regions for every scene query. Concretely, let $X = (x,y,z,1)^\top$ denote a 3D point in the LiDAR/ego frame associated with $q_{ij}$ and height bin $z$, For camera i with pose $[R_i|t_i]$ and intrinsics $K_i$ , the 3D-to-2D projection is 
\begin{align}
\tilde{\mathbf{u}}_i = K_i\,[\,R_i\mid t_i\,]\,X, \quad u_i = \frac{\tilde{u}_{i,0}}{\tilde{u}_{i,2}}, v_i = \frac{\tilde{u}_{i,1}}{\tilde{u}_{i,2}},
\end{align}
and we aggregate features at sub-pixel locations via bilinear sampling $\phi(F_i, u_i, v_i)$. The scene representation for cell $q_{ij}$ is then obtained by deformable cross-attention over $K$ learned offsets per camera:
\begin{align}
    \mathbf{b}_{ij} = \sum_{i=1}^{N}\sum_{k=1}^{K}\alpha^{ij}_{i,k} \phi(F_i, u^{ij}_{i,k}, v^{ij}_{i,k}),
\end{align}
where $\alpha^{ij}_{i,k}$ are attention weights produced by scene encoder and depend on the query/content features and geometry. 

Because the scene grid size $(H_b, W_b)$ is fixed (e.g., $H_b$=128, $W_b$=128), the number of resulting scene tokens $N_b = H_b \times W_b$ is decoupled from both the number of cameras $N$ and the input resolution $H \times W$, yielding a consistent token budget across rigs and image sizes. Finally, we enrich scene queries with positional encodings over the physical range to encode metric coordinates. The result is the geometry-aware scene tokens $B \in \mathbb{R}^{H_b \times W_b \times C_b}$ that preserve high-level semantics from the foundation backbone while remaining camera- and resolution-agnostic, ready to be consumed by our spatial-aware multi-view transformer.

\vspace{-1mm}
\subsection{Spatial-Aware Multi-View Decoder}
To enable structured interaction between the unified scene tokens and per-camera view tokens, we introduce the spatial-aware multi-view transformer. Unlike conventional dense cross-attention that treats all token pairs equally, our transformer employs the visibility constraint to ensure that only physically visible scene regions interact with the corresponding camera views.

Specifically, our spatial-aware multi-view transformer establishes visibility-guided scene-view attention between the scene tokens and the view tokens within a standard ViT~\cite{dosovitskiy2020image} framework. Each scene token represents an ego-centric spatial cell, while each view token is initialized from its 2D positional embedding $\mathbf{E}$ plus a Plücker embedding $\mathbf{P}$ of the corresponding viewing ray in the ego frame: 
\begin{align}
\mathbf{v}_{i,p} = \mathbf{E}^{\mathrm{2D}}_{i,p}
 \;+\; \mathbf{P}\!\big(\ell_{i,p}\big),
\end{align}
where $i$ indexes the camera and $p$ indexes the image patch. The plücker ray $\ell_{i,p} = (\hat{d},m)$ encodes the ray that starts at the camera center and passes through patch $p$:
\begin{align}
\ell_{i,p} = \text{PlückerRay}\big(K_i, R_i, t_i, \tilde{\mathbf u}_{i,p}\big),\\
\mathbf{P}\!\big(\ell_{i,p}\big) = \mathrm{MLP}\big([\hat{\mathbf d}_{i,p};\,\mathbf m_{i,p}]\big).
\end{align}
By injecting $\ell_{i,p}$, we distinguish tokens from different viewpoints that may share similar 2D appearance and provides a camera-aware geometric prior, which improves cross-view disambiguation and alignment with the 3D space.

To ensure physically meaningful communication between these two token types, we introduce a visibility-based attention mask $M$ that masks out impossible scene-view correspondences. Let the scene tokens be 
\begin{align}
    \mathbf{B} = [\mathbf{b}_1,\mathbf{b}_2,\ldots,\mathbf{b}_{N_b}] \in \mathbb{R}^{N_b \times D},
\end{align}
and the concatenated view tokens from all cameras be
\begin{align}
    \mathbf{V} = [\mathbf{v}_{1,1},\ldots,\mathbf{v}_{i,p},\ldots,\mathbf{v}_{N,N_p}] \in \mathbb{R}^{N_v \times D},
\end{align}
where $N$ is the number of camera views, $N_p = H_p \times W_p$ is the number of image patches per view, $N_b$ and $N_v$ denote the total numbers of scene and view tokens, respectively.

For each camera $i$ and scene cell $j$, we precompute a binary visibility mask $M$:
\begin{align}
    M_{i,j} =
        \begin{cases}
        1, & \text{if scene cell } j \text{ is visible in camera } i, \\
        0, & \text{otherwise.}
        \end{cases}
\end{align}
This mask is applied symmetrically to both view$\rightarrow$scene and scene$\rightarrow$view directions, while the scene$\rightarrow$scene and view$\rightarrow$view directions remain unmasked.

Within each ViT block, the visibility-aware multi-head self-attention is computed as
\begin{align}
\mathrm{Attn}(\mathbf{Q},\mathbf{K},\mathbf{V},\mathbf{M}) = \mathrm{softmax}\!\left(\frac{\mathbf{Q}\mathbf{K}^\top}{\sqrt{d}}\odot \mathbf{M} \right)\mathbf{V}.
\end{align} 

After passing through $L$ stacked transformer layers, the updated token sequence becomes
\begin{align}
[\mathbf{B}',\,\mathbf{V}']
  = \mathrm{Transformer}_L([\mathbf{B},\,\mathbf{V}],\,\mathbf{M}).
\end{align}

This architecture enables bi-directional, geometry-consistent information exchange: scene tokens aggregate view-specific details from the cameras that observe them, while view tokens are enriched with scene-level spatial context reflecting the true 3D layout of the driving scene. Consequently, the resulting unified scene tokens maintain both semantic and geometric consistency across all camera views, forming a coherent spatial foundation for downstream perception and reasoning tasks.

\vspace{-1mm}
\subsection{Unified Reconstruction and Understanding}
We utilize a joint task training strategy to enrich the unified scene tokens with texture, semantics, and 3D geometry. These unified scene tokens provide VLMs with dense and well-structured perceptual input for safer and more accurate decision-making. We adopt three image-space tasks: image reconstruction, depth prediction, and semantic prediction, as well as an occupancy prediction task and a semantic regularization for scene tokens. The image space tasks utilize a DPT~\cite{ranftl2021vision} decoder that upsamples per camera view tokens to the native image resolution, while the occupancy prediction task employs a convolutional head to generate occupancy logits from scene tokens.
The semantic regularization term explicitly injects semantic information into the scene tokens, promoting semantic structure learning in the latent space.

\textbf{Image Reconstruction.} Given the updated view tokens $V$, the DPT head predicts per camera RGB images $\hat{\mathbf{I}}$. The reconstruction loss combines pixel-level fidelity, perceptual similarity, and adversarial training:
\begin{align}
    \mathcal{L}_{\mathrm{rgb}} = \lambda_{\mathrm{pix}} \|\hat{\mathbf{I}} - \mathbf{I}\|_1 + \lambda_{\mathrm{perc}} \mathcal{L}_{\mathrm{LPIPS}} + \lambda_{\mathrm{adv}} \mathcal{L}_{\mathrm{GAN}},
\end{align}
where $\|\cdot\|_1$ denotes the L1 distance evaluated over valid pixels, $\mathcal{L}_{\mathrm{LPIPS}}$ is the learned perceptual image patch similarity loss, and $\mathcal{L}_{\mathrm{GAN}}$ is the adversarial loss computed via a discriminator following~\cite{esser2021taming}. In our implementation, we set the balancing weights to $\lambda_{\mathrm{pix}} = 1.0$, $\lambda_{\mathrm{perc}} = 1.0$, and $\lambda_{\mathrm{adv}} = 0.3$.

\textbf{Depth Prediction.} Public driving datasets~\cite{caesar2020nuscenes, geiger2013vision} rarely provide dense per-pixel depth ground truth. The sparsity limits pixel-wise supervision, encourages texture copying artifacts, and leaves monocular depth with unresolved metric scale, which is known to bottleneck downstream perception and planning quality. We first obtain dense pseudo depth for each pixel using MoGe-2~\cite{wang2025moge2}, which predicts metric-scale monocular geometry. We then project these LiDAR points onto each image plane to obtain sparse metric samples that anchor the pseudo labels. To reconcile the two, we estimate a global scale and shift per image using ROE~\cite{wang2025moge}, which down-weights outliers from occlusions and sensor noise and produces stable alignment. The aligned pseudo depth $d^{align}$ serves as the supervision target for a DPT-style depth head. The depth head predicts $\hat{d}$ and is trained with a Charbonnier term and a gradient consistency regularizer that preserves sharp boundaries:
\begin{equation}
\begin{split}
\mathcal{L}_{\mathrm{depth}}
= \sqrt{(\hat{d}-d^{\mathrm{align}})^2+\epsilon^2} 
\quad + \gamma_{\nabla}\sum_{a\in\{x,y\}}
   \sqrt{(\nabla_a \hat{d}-\nabla_a d^{\mathrm{align}})^2+\epsilon^2},
\end{split}
\end{equation}
where $\epsilon = 0.001$ is the Charbonnier constant, $\gamma_{\nabla}$ is the weight of the gradient-consistency term, $\nabla_a$ denotes the discrete image gradient along direction $a \in \{x, y\}$, and the summation aggregates horizontal and vertical components.

\textbf{Semantic Prediction.} We obtain semantic supervision by projecting LiDARSeg labels onto the image plane, which produces sparse yet precise per pixel ground truth. The DPT decoder outputs per pixel logits and we use cross-entropy with an ignore mask so that only pixels covered by the LiDARSeg projection contribute to the loss:
\begin{align}
\mathcal{L}_{\mathrm{sem}} = \mathrm{CE}(\hat{S}, Y; \mathrm{ignore\_label} = 255),
\end{align}
where $\hat{S}$ are the logits from decoder, $Y$ are sparse projected labels, and ignore label masks pixels without supervision.

\textbf{Occupancy Prediction.} To further enhance the geometric and semantic expressiveness of our unified scene representation, we incorporate an additional occupancy prediction objective~\cite{huang2023tri} during training. This task complements the image-space objectives by encouraging the scene tokens to reason about the 3D structure of the driving environment. Specifically, the scene tokens $B$ are decoded by a 3D occupancy head $\mathcal{D}_{occ}$ to predict voxel-wise semantic occupancy logits $\hat{O} \in \mathbb{R}^{X \times Y \times Z \times C}$.
The occupancy prediction is optimized with a combination of the cross-entropy loss and the Lovász-Softmax loss, formulated as: 
\begin{align} 
    \mathcal{L}_{\text{occ}} = \mathcal{L}_{\text{CE}}(\hat{O}, O) + \lambda_{\text{Lovász}}\mathcal{L}_{\text{Lovász}}(\hat{O}, O), 
    \label{eq:occ}
\end{align} 
where $O$ denotes the ground-truth occupancy labels, $\lambda_{\text{Lovász}}$ is set to 0.2. Jointly optimizing the occupancy and image-space tasks enables the scene tokens to capture richer 3D spatial and semantic relationships, thereby facilitating more holistic scene understanding.

\textbf{Semantic Regularization.}
Conventional perception methods in autonomous driving apply supervision only on the model outputs, neglecting the semantic structure of the underlying scene representation.
Apart from image-space and 3D occupancy supervisions, we further regularize the scene tokens by aligning them with explicit semantic information to avoid structure corruption in the latent space.
To elaborate, we first derive the ground truth semantic labels for scene tokens by projecting semantic occupancy labels to the latent space with importance weighting for different semantic classes.
Then we regularize the scene tokens with $\mathcal{L}_{\mathrm{reg}}$ consisting of the cross entropy loss and the Lovász-Softmax loss, which shares the same form as Eq.~(\ref{eq:occ}) but in the latent space of scene tokens.

\vspace{-3mm}
\paragraph{Total Training Objective.}
The overall training loss is defined as:
\vspace{-3mm}
\begin{equation}
\begin{split}
\mathcal{L}_{\mathrm{total}}=\lambda_{\mathrm{rgb}}\,\mathcal{L}_{\mathrm{rgb}}\,+\,\lambda_{\mathrm{depth}}\,\mathcal{L}_{\mathrm{depth}} 
\,+\,\lambda_{\mathrm{sem}}\,\mathcal{L}_{\mathrm{sem}} \,+\, \lambda_{\mathrm{occ}}\,\mathcal{L}_{\mathrm{occ}}\,+\, \lambda_{\mathrm{reg}}\,\mathcal{L}_{\mathrm{reg}}.
\end{split}
\end{equation}
We set $\lambda_{\mathrm{rgb}}$ to 10.0 and $\lambda_{\mathrm{occ}}$ to 5.0, while $\lambda_{\mathrm{depth}}$ to 0.2, $\lambda_{\mathrm{sem}}$ to 0.1 and $\lambda_{\mathrm{reg}}$ to 3.0.

We think the main contribution of our method is the formulation of this unified scene tokenizer, paving the way for new generative paradigms for autonomous driving. 
For some components (e.g., BEV queries), we employ simple yet effective designs to instantiate our formulation. We adopt new designs (e.g., visibility-aware attention) when existing techniques cannot well fit.

%% file: sec/4_experiment.tex
\begin{table}[t]
  \centering
  \small
  \setlength{\tabcolsep}{2pt}
\caption{\textbf{Comparison of image tokenizers on nuScenes dataset.} $\ddagger$ denotes VQGAN was trained on Imagenet~\cite{deng2009imagenet}, $\dagger$ indicates trained on OpenImages~\cite{kuznetsova2020open}. Tok. denotes the tokenization stage only (w/o decoder).}
  \label{tab:cvpr_psnr_ssim}
     \vspace{-3mm}
\resizebox{\linewidth}{!}{
  \begin{tabular}{lccccc}
    \toprule
    Method & Input 
    & PSNR $\uparrow$ & SSIM $\uparrow$
    & Latency (Tok./Full) $\downarrow$
    & Peak Memory (Tok./Full) $\downarrow$ \\
    \midrule
    $\text{VQGAN}^{\ddagger}$~\cite{esser2021taming} 
      & $256\times256$ & 24.41 & 0.674 & 63.31 ms / \bf{77.06 ms} & 1440.7 MB / 1608.9 MB \\
    $\text{ViT-VQGAN}$~\cite{yu2021vector}
      & $256\times256$ & 26.96 & 0.763 & 59.23 ms / 117.36 ms & \bf{1354.4 MB} / \bf{1445.1 MB} \\
    $\text{VQGAN}^{\dagger}$~\cite{esser2021taming} 
      & $256\times256$ & 26.77 & 0.756 & 63.34 ms / 115.17 ms & 1440.7 MB / 1605.8 MB \\
    $\text{FlowMo-Lo}$~\cite{sargent2025flow}
      & $256\times256$ & 23.11 & 0.696 & 248.04 ms / 5555.78 ms & 8434.8 MB / 8464.5 MB \\
    $\text{FlowMo-Hi}$~\cite{sargent2025flow}
      & $256\times256$ & \textbf{27.91} & \textbf{0.793} & 259.70 ms / 5562.63 ms & 8436.6 MB / 8466.6 MB \\
      \midrule
      BEV-VAE~\cite{chen2025bev}  & $ 6\times256\times256$ & 26.32  & 0.746 & -- / -- & -- / --\\
    \textbf{DriveTok} 
      & $6\times256\times256$ & 27.13 & 0.723 & \textbf{21.86 ms} /  267.82 ms & 3957.95 MB / 7921.09 MB \\
    \textbf{DriveTok} & $6\times256\times704$ & 27.89 & 0.747 & 87.96 ms / 382.51 ms & 4423.73 MB / 8456.09 MB \\
    \bottomrule
  \end{tabular}
  }
  \label{tab:image}  
\vspace{-5mm}
\end{table}

\section{Experiments}
\label{sec:exp}
\subsection{Implementation Details}
\paragraph{Dataset and Inputs.} We train DriveTok on nuScenes~\cite{caesar2020nuscenes} dataset using 6 surrounding cameras. All images are resized to 256 $\times$ 704, and these settings are shared across training and evaluation. For supervision, depth is produced per pixel by MoGe-2~\cite{wang2025moge2} and then aligned to metric scale via a robust ROE scale–shift fit using sparse LiDAR depths projected to the image plane. The aligned depth serves as the training target. For semantic labels, we project nuScenes-lidarseg annotations onto each camera to obtain sparse yet precise per-pixel supervision, with an $\text{ignore\_mask = 255}$ applied to unlabeled pixels. Additionally, we employ semantic occupancy labels from SurroundOcc~\cite{wei2023surroundocc} for supervision on occupancy prediction task.

\textbf{Model and Training Setup.} The pretrained visual foundation backbone is a pretrained DINOv3-ViTB~\cite{simeoni2025dinov3} with patch size 16, producing a 4-level feature pyramid with channel widths $\{768, 768, 768, 768\}$. A 4-level FPN fuses these features and outputs a $C=768$ feature map that is lifted by a BEVFormer-style module to a 128 $\times$ 128 BEV grid ($C_b=768$). The BEV region of interest is fixed to $\text{pc\_range} = [-51.2m,\,-51.2m,\,-5.0m,\,51.2m,\,51.2m,\,3.0m]$ in the lidar coordinate frame, and we use 8 vertical bins for visibility checks. The spatial-aware multi-view transformer follows a ViT-Base configuration for scene $\leftrightarrow$ view token interaction. All image-space task heads adopt a DPT-style decoder that converts tokens to dense pixels. Moreover, we employ a convolution-based occupancy head that lifts BEV features into 3D voxels for occupancy prediction. The total number of trainable parameters is approximately 280M.

For optimization, we use AdamW~\cite{loshchilov2017decoupled} with learning rate $1 \times 10^{-4}$, weight decay 0.01, $\beta = (0.9,0.999)$. Global gradient clipping is set to 35.0. A cosine learning-rate schedule with warmup drives updates in steps using the exact number of iterations per epoch. To accelerate training and reduce consumption, we enable FlashAttention-2~\cite{dao2023flashattention} and train in BFloat16 precision. The model is trained for about 400k iterations on 8× A800 GPUs.

\begin{table}[t]
  \centering
  \small
  \setlength{\tabcolsep}{8pt}
  \caption{\textbf{Comparison of different monocular depth estimator on nuScenes dataset.} We utilize the projected lidar points as ground-truth.}
     \vspace{-3mm}
  \label{tab:cvpr_absrel}
  \resizebox{\linewidth}{!}{
  \begin{tabular}{lccccc}
    \toprule
    Method & Input & absrel $\downarrow$ & $\delta<1.25$ $\uparrow$ & latency $\downarrow$ &  Peak Memory $\downarrow$ \\
    \midrule
    UniDepthV2-B~\cite{piccinelli2025unidepthv2} & $256\times704$ & 0.52 & 0.30 & \bf62.45 ms & \bf2194.65 MB \\
    UniDepthV2-L~\cite{piccinelli2025unidepthv2} & $ 256\times704$ & 0.40 & 0.31 & 100.45 ms & 3902.32 MB\\
    DepthPro~\cite{bochkovskii2024depth} & $256\times704$ & 0.26 & 0.47 & 5179.93 ms & 30559.55 MB \\
    Metric3D-V2~\cite{hu2024metric3d} & $256\times704$ & 0.28 & 0.69 & 403.49 ms & 2982.53 MB \\
    \midrule
    \textbf{DriveTok} & $6\times256\times704$ & \bf0.08 & \bf0.93 &383.53 ms & 8562.64 MB \\
    \bottomrule
  \end{tabular}
    \label{tab:depth}
  }
     \vspace{-6mm}
\end{table}

\textbf{Evaluation and Metrics.} All tasks are evaluated on the nuScenes~\cite{caesar2020nuscenes} validation set. For image reconstruction task, we use Peak Signal-to-Noise Ratio (PSNR) and Structural Similarity Index (SSIM) as quantitative measures of reconstruction quality. For depth prediction task, We utilize the aligned dense depth as ground-truth, and we report the absolute relative error (AbsRel) and the percentage of pixels satisfying $\delta<1.25$ as evaluation metrics. 

For semantic prediction task, since most existing methods employ dense pseudo-labels generated by segmentation models such as SAM~\cite{kirillov2023segment}, while our goal is to enrich the scene tokens with semantic awareness rather than achieve dense segmentation accuracy, we supervise the model only with sparse LiDARSeg ground truth. To verify that our unified scene tokens encode meaningful semantic information, we provide qualitative visualization results in \cref{sec:sem}.

Finally, for the 3D occupancy prediction task, evaluation is conducted on the SurroundOcc using the same metrics as in~\cite{huang2024gaussianformer}, namely mIoU and IoU, to measure both semantic and geometric prediction quality.

\vspace{-1mm}
\subsection{Image Reconstruction}
We compared DriveTok with representative image tokenizers to assess whether our method encodes multi-view inputs into unified scene tokens that preserve fine-grained appearance information. As shown in \cref{tab:image}, our method achieves performance comparable to all baselines across all six cameras, demonstrating that DriveTok can be adapted to multi-view inputs in autonomous driving. Additionally, we present qualitative results of our method in \cref{fig:vis1}. DriveTok reconstructs textures and preserves cross-view consistency at overlapping fields of view. These observations highlight the advantage of adopting a unified scene representation for surround-view scenarios.

\begin{table}[t] \small
  \centering
  \footnotesize
  \setlength{\tabcolsep}{2pt}
  \caption{\textbf{Comparison of different multi-view depth predictors on the nuScenes dataset.} We utilize the projected lidar points as ground-truth.}
  \vspace{-2mm}
  \label{tab:depth2}
  \begin{tabular}{lcc|@{\hspace{8pt}}lcc}
    \toprule
    Method & absrel $\downarrow$ & $\delta<1.25$ $\uparrow$ & Method & absrel $\downarrow$ & $\delta<1.25$ $\uparrow$ \\
    \midrule
    SurroundDepth~\cite{wei2023surrounddepth} & 0.28 & 0.66 & Dist4D~\cite{guo2025dist}   & 0.39 & 0.58 \\
    R3D3~\cite{schmied2023r3d3}          & 0.25 & 0.73 & OmniNWM~\cite{li2025omninwm}  & 0.23 & 0.81 \\
    SelfOcc~\cite{huang2024selfocc}       & 0.23 & 0.75 & \textbf{DriveTok} & \textbf{0.08} & \textbf{0.93} \\
    \bottomrule
  \end{tabular}
  \vspace{-2mm}
\end{table}

\vspace{-1mm}
\subsection{Depth Prediction}
We evaluated the effectiveness of DriveTok in learning depth-aware representations that capture both metric scale and scene geometry from multi-view inputs. We compared DriveTok with representative monocular depth estimation methods and multi-view depth predictors. As shown in \cref{tab:depth} and \cref{tab:depth2}, DriveTok achieves the lowest AbsRel and the highest $\delta<1.25$ among all compared methods. These results demonstrate that the unified scene tokens effectively capture both local geometry and global scene structure, forming a reliable geometric foundation for downstream perception tasks. Moreover, we further project the predicted surround-view depth into the world coordinate system to obtain 3D reconstructed point clouds, as illustrated in \cref{fig:vis1}, which further verifies the superiority of our method in maintaining geometry consistency.

\begin{table*}[t] %
    \caption{\textbf{3D semantic occupancy prediction results on nuScenes.}}

    \small
    \setlength{\tabcolsep}{0.001\linewidth}  
    \vspace{-3mm}  
    \renewcommand\arraystretch{1.1}
    \centering
    \resizebox{\textwidth}{!}{
    \begin{tabular}{l|c c | c c c c c c c c c c c c c c c c}
        \toprule
        Method
        & IoU
        & mIoU
        & \rotatebox{90}{\textcolor{nbarrier}{$\blacksquare$} barrier}
        & \rotatebox{90}{\textcolor{nbicycle}{$\blacksquare$} bicycle}
        & \rotatebox{90}{\textcolor{nbus}{$\blacksquare$} bus}
        & \rotatebox{90}{\textcolor{ncar}{$\blacksquare$} car}
        & \rotatebox{90}{\textcolor{nconstruct}{$\blacksquare$} const. veh.}
        & \rotatebox{90}{\textcolor{nmotor}{$\blacksquare$} motorcycle}
        & \rotatebox{90}{\textcolor{npedestrian}{$\blacksquare$} pedestrian}
        & \rotatebox{90}{\textcolor{ntraffic}{$\blacksquare$} traffic cone}
        & \rotatebox{90}{\textcolor{ntrailer}{$\blacksquare$} trailer}
        & \rotatebox{90}{\textcolor{ntruck}{$\blacksquare$} truck}
        & \rotatebox{90}{\textcolor{ndriveable}{$\blacksquare$} drive. suf.}
        & \rotatebox{90}{\textcolor{nother}{$\blacksquare$} other flat}
        & \rotatebox{90}{\textcolor{nsidewalk}{$\blacksquare$} sidewalk}
        & \rotatebox{90}{\textcolor{nterrain}{$\blacksquare$} terrain}
        & \rotatebox{90}{\textcolor{nmanmade}{$\blacksquare$} manmade}
        & \rotatebox{90}{\textcolor{nvegetation}{$\blacksquare$} vegetation}
        \\
        \midrule
        MonoScene~\cite{cao2022monoscene} & 23.96 & 7.31 & 4.03 &	0.35& 8.00& 8.04&	2.90& 0.28& 1.16&	0.67&	4.01& 4.35&	27.72&	5.20& 15.13&	11.29&	9.03&	14.86 \\
        
        Atlas~\cite{atlas} & 28.66 & 15.00 & 10.64&	5.68&	19.66& 24.94& 8.90&	8.84&	6.47& 3.28&	10.42&	16.21&	34.86&	15.46&	21.89&	20.95&	11.21&	20.54 \\
        
        BEVFormer~\cite{li2024bevformer} & 30.50 & 16.75 & 14.22 &	6.58 & 23.46 & 28.28& 8.66 &10.77& 6.64& 4.05& 11.20&	17.78 & 37.28 & 18.00 & 22.88 & 22.17 & {13.80} &	22.21\\

        TPVFormer~\cite{huang2023tri}  & {30.86} & 17.10 & 15.96&	 5.31& 23.86	& 27.32 & 9.79 & 8.74 & 7.09 & 5.20& 10.97 & 19.22 & {38.87} & {21.25} & {24.26} & {23.15} & 11.73 & 20.81\\

        OccFormer~\cite{zhang2023occformer} & {31.39} & {19.03} & {18.65} & {10.41} & {23.92} & {30.29} & {10.31} & {14.19} & {13.59} & {10.13} & {12.49} & {20.77} & {38.78} & 19.79 & 24.19 & 22.21 & {13.48} & {21.35}\\
        
        SurroundOcc~\cite{wei2023surroundocc} & {31.49} & {20.30}  & {20.59} & {11.68} & \textbf{28.06} & \textbf{30.86} & {10.70} & {15.14} & \textbf{14.09} & \textbf{12.06} & \textbf{14.38} & \textbf{22.26} & 37.29 & {23.70} & {24.49} & {22.77} & 14.89 & {21.86}  \\
        
        GaussianFormer~\cite{huang2024gaussianformer} & 29.83 & {19.10} & {19.52} & {11.26} & {26.11} & {29.78} & {10.47} & {13.83} & {12.58} & {8.67} & {12.74} & {21.57} & {39.63} & {23.28} & {24.46} & {22.99} & 9.59 & 19.12 \\

        GaussianFormer-2~\cite{huang2025gaussianformer2} & 30.56 & {20.02} & \textbf{20.15} & {12.99} & {27.61} & {30.23} & {11.19} & {15.31} & {12.64} & {9.63} & {13.31} & \textbf{22.26} & {39.68} & {23.47} & {25.62} & {23.20} & 12.25 & 20.73 \\

        QuadricFormer~\cite{zuo2025quadricformer} &31.22 & \textbf{20.12} & 19.58 & \textbf{13.11} & 27.27 & 29.64 & \textbf{11.25} & \textbf{16.26} & 12.65 & 9.15 & 12.51 & 21.24 & 40.20 & 24.34 & 25.69 & 24.24 & 12.95 & 21.86 \\
       
        \textbf{DriveTok} & \textbf{33.32} & 20.06 & 20.04 & 9.81 & 26.11 & 27.87 & 11.14 & 13.44 & 9.52 & 8.65 & 12.67 & 20.44 & \bf41.79 & \bf25.31 & \bf27.68 & \bf26.37 & \bf16.59 & \bf23.57 \\

        \bottomrule
    \end{tabular}}
    \label{tab:nuscenes_results}
    \vspace{-7mm}
\end{table*}

\vspace{-1mm}
\subsection{Occupancy Prediction}
We further evaluated our DriveTok by testing its capability on occupancy prediction task, which directly measures how well the learned tokens capture geometric and semantic structure of the driving scene. Specifically, We compared our DriveTok against recent occupancy predition models, including BEVFormer~\cite{huang2021bevdet}, GaussianFormer~\cite{huang2024gaussianformer} and Quadricformer~\cite{zuo2025quadricformer}. The quantitative results summarized in \cref{tab:nuscenes_results} show that our DriveTok achieves performance comparable to state-of-the-art methods under the same setting, highlighting its superior ability to learn rich spatial and semantic information from multi-view inputs.

\vspace{-3mm}
\subsection{Semantic Prediction}
\label{sec:sem}
\begin{figure*}[t]
     \centering
     \includegraphics[width=\linewidth]{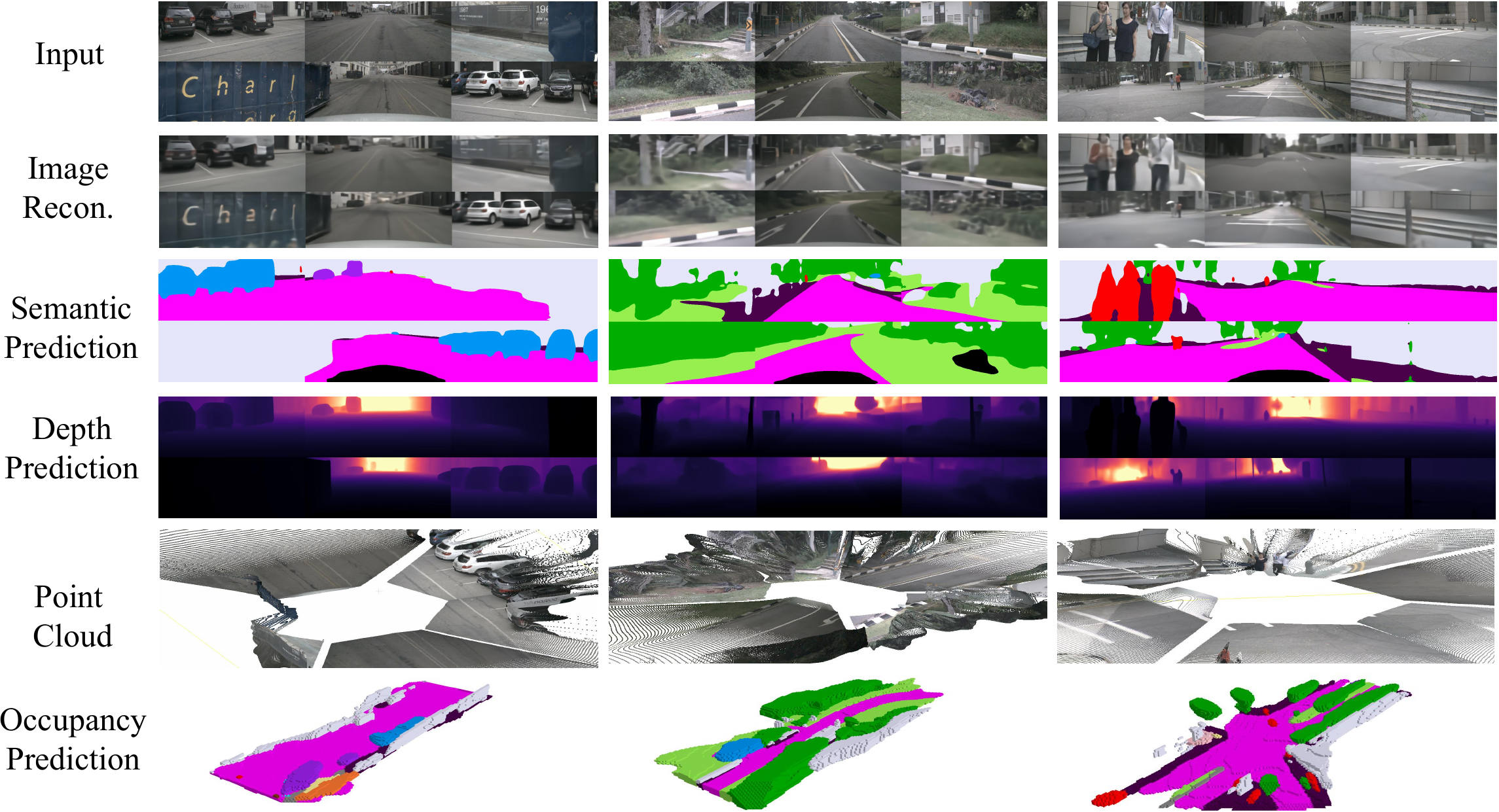}
     \vspace{-7mm}
     \captionof{figure}{\textbf{Visualizations of DriveTok in different tasks.} We provide a holistic visualization of our DriveTok in diverse autonomous driving scene reconstruction and understanding tasks. They show that DriveTok effectively represents the overall 3D environment by constructing scene tokens, thereby maintaining strong multi-view consistency. Through joint task training, the learned scene tokens not only capture texture details in the image space but also achieve a deeper perception and understanding.}
 \label{fig:vis1}
     \vspace{-2mm}
 \end{figure*}

\vspace{-2mm}
To further verify whether our scene tokens encode semantic information, we conducted qualitative visualizations on the semantic prediction task, as shown in \cref{fig:vis1}.
Although trained only with sparse LiDARSeg labels, DriveTok successfully predicts coherent semantic layouts that align well with object semantics and scene geometry. The predictions remain consistent across adjacent viewpoints, indicating that our scene tokens capture strong cross-view semantic correspondences. The model also preserves clear object boundaries even in cluttered environments. Moreover, DriveTok reliably differentiates complex scene elements such as sidewalks, vegetation, and distant structures.

\begin{figure*}[t]
     \centering
     \includegraphics[width=\linewidth]{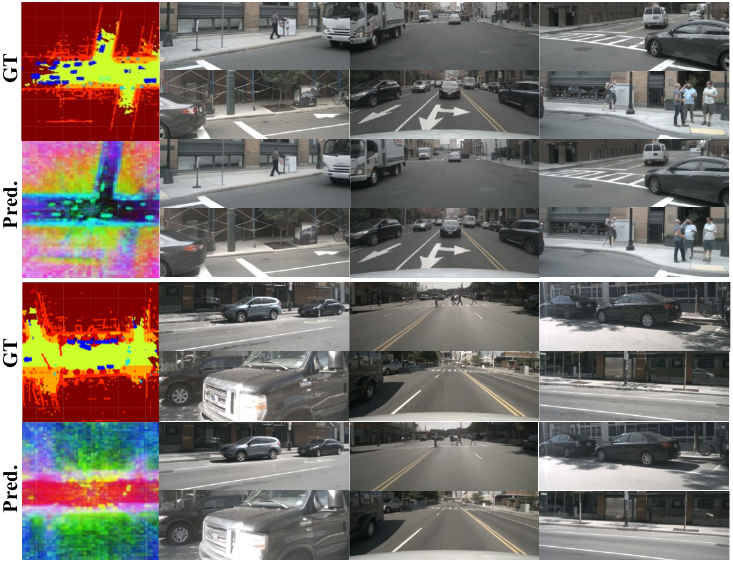}
     \vspace{-7mm}
     \captionof{figure}{\textbf{Visualizations of BEV scene tokens and images.}
     We visualize the BEV feature maps with PCA, the ground truth labels for semantic regularization and the decoded images v.s. ground truth images.
     The PCA result clearly shows that our BEV scene tokens not only learns the complex textures but also models the semantic structure of the driving scenes, avoiding radial patterns in conventional methods.
     }
 \label{fig:vis bev}
     \vspace{-7mm}
 \end{figure*}%

\subsection{Ablation Study}
\textbf{Visualizations.}
We visualize the BEV feature maps with principle component analysis, the groundtruth labels for semantic regularization and the decoded images v.s. ground truth images in Fig.~\ref{fig:vis bev}.
The proposed semantic regularization successfully injects explicit semantic structure into the BEV latent space, avoiding the radial patterns common in conventional perception methods.

\textbf{Visibility-Guided Attention.} To assess the effectiveness of the visibility-guided attention in our spatially aware multi-view decoder, we conducted an ablation study in which we removed the visibility mask and allowed each scene token to attend to all view tokens unconditionally. Specifically, we adopted a two-stage training protocol. We first pretrained the scene tokens using only the image reconstruction task. After pretraining, we froze all components except the occupancy head and fine-tuned solely on the 3D occupancy prediction task to evaluate whether the visibility-guided attention forces the scene tokens to encode geometric information. As illustrated in \cref{tab:ablation1}, removing visibility-guided attention eliminates the local geometric inductive bias, causing scene tokens to overfit image textures rather than learn meaningful spatial structure. Such texture-biased tokens fail to support reliable 3D understanding in autonomous driving. In contrast, visibility-guided attention serves as an effective regularizer that prevents overfitting and enforces physically valid interactions, thereby encouraging the model to encode spatial information.

\begin{table}[t] \small
    \centering
    \small
    \setlength{\tabcolsep}{10pt}
    \caption{\textbf{Ablation on visibility-guided attention.} Removing visibility guidance degrades spatial reasoning, causing scene tokens to overfit image textures.}
    \vspace{-3mm}
    \label{tab:vis_attn_ablation}
    \begin{tabular}{l|cc|cc}
        \toprule
        \multirow{2}{*}{Setting} 
            & \multicolumn{2}{c|}{Texture} 
            & \multicolumn{2}{c}{Geometry \& Semantics} \\
        \cmidrule(lr){2-3} \cmidrule(lr){4-5}
            & PSNR$\uparrow$ & SSIM$\uparrow$ 
            & \phantom{xx}IoU$\uparrow$ & \phantom{xx}mIoU$\uparrow$ \\
        \midrule
        w/o Geo-Attn & \bf29.98 & \bf0.798 & \phantom{xx}5.32 & \phantom{xx}0.59 \\
        \textbf{w/ Geo-Attn} & 28.84 & 0.770 & \phantom{xx}\textbf{12.81} & \phantom{xx}\textbf{3.84} \\
        \bottomrule
    \end{tabular}
    \label{tab:ablation1}
    \vspace{-4mm}
\end{table}

\begin{table*}[t] \small
    \centering
    \small
    \setlength{\tabcolsep}{2pt}
    \caption{\textbf{Ablation study on the joint task training strategy.} Different combinations of reconstruction, depth, semantic, and occupancy objectives reveal a trade-off between texture quality and geometry-semantic understanding, where stronger multi-task supervision improves occupancy prediction performance.}

    \vspace{-3mm}
    \label{tab:ablation_joint_training}
    \resizebox{\textwidth}{!}{
        \begin{tabular}{l|cccc|cc|cc}
            \toprule
            \multirow{2}{*}{Setting} 
                & \multicolumn{4}{c|}{Tasks} 
                & \multicolumn{2}{c|}{Texture} 
                & \multicolumn{2}{c}{Geometry \& Semantics} \\
            \cmidrule(lr){2-5} \cmidrule(lr){6-7} \cmidrule(lr){8-9}
                & Recon. & Depth & Sem. & Occ. 
                & PSNR$\uparrow$ & SSIM$\uparrow$ 
                & \phantom{xx}IoU$\uparrow$ & \phantom{xx}mIoU$\uparrow$ \\
            \midrule
            Image Tokenizer               & \checkmark &            &           &           & \bf28.84 & \bf0.770 & \phantom{xx}12.81 & \phantom{xx}3.84 \\
            Geometry-Aware Model             & \checkmark & \checkmark &           &           & 28.52 & 0.764 & \phantom{xx}20.72 & \phantom{xx}9.36 \\
            Scene-Understanding Model        & \checkmark & \checkmark & \checkmark &          & 28.06 & 0.756 & \phantom{xx}24.05 & \phantom{xx}11.68 \\
            Unified Model   & \checkmark & \checkmark & \checkmark & \checkmark & 27.89 & 0.747 & \phantom{xx}\textbf{33.32} & \phantom{xx}\textbf{20.06} \\
            \bottomrule
        \end{tabular}
        \label{tab:ablation2}
    }
    \vspace{-8mm}
\end{table*}

\textbf{Joint Task Training Strategy.} To validate the effectiveness of our joint task training strategy, we trained the scene tokens under different combinations of training objectives and then fine-tuned the occupancy head and measured the performance of occupancy prediction. The results in \cref{tab:ablation2} show that the image reconstruction task encourages the scene tokens to retain low-level texture information from the input images, while the depth and semantic prediction tasks enable the scene tokens to capture rich 3D spatial structure and semantic cues, allowing the model to better perceive and understand the surrounding environment and thus support 3D reasoning for autonomous driving. Finally, although incorporating the occupancy prediction task degrades the image reconstruction quality, it further enhances the spatial and semantic information encoded in the scene tokens, a trade-off that we consider worthwhile.

%% file: sec/5_conclusion.tex
\section{Conclusion and Discussions}
In this work, we introduced DriveTok, a unified scene tokenizer for multi-view driving scene reconstruction and understanding. DriveTok produces compact scene tokens that simultaneously encode texture, geometry, and semantics via the joint task training strategy. Our spatial-aware multi-view decoder, equipped with visibility-guided attention, enables geometry-consistent interaction between scene tokens and visual tokens. Extensive experiments on nuScenes~\cite{caesar2020nuscenes} demonstrate that DriveTok achieves competitive image reconstruction quality and state-of-the-art 3D semantic occupancy performance under the same frozen-backbone setting. Moreover, DriveTok attains the best results on the metric depth prediction task, validating the effectiveness of our tokenization framework.

\textbf{Future Directions}. We envision DriveTok as a general visual interface for vision-language(-action) models and driving world models. The unified scene tokens offer a compact yet semantically rich representation that can be readily consumed by large models for downstream tasks such as open-ended driving question answering, counterfactual reasoning, multi-step planning, and video prediction. Integrating DriveTok with VLAs and world models could enable more scalable training and facilitate closed-loop policy learning where high-level reasoning and low-level perception share a common spatial memory. Moreover, extending our framework to incorporate temporal modeling, additional sensor modalities (e.g., LiDAR or radar), and generative capabilities (e.g., scene editing or future synthesis) is a promising direction for building holistic driving foundation models.

%% file: sec/X_suppl.tex
\appendix
\section{Architecture and Implementation Details}
\label{sec:archi}
This section provides more implementation details of the proposed architecture, including 3D scene encoder, spatial-aware multi-view decoder, task-specific heads, and the implementation of our visibility-guided scene-view attention. 
\vspace{-3mm}
\subsection{3D Scene Encoder}
\textbf{Visual Foundation Model and Feature Sizes.} The 3D scene encoder is responsible for transforming multi-view image observations into unified scene tokens suitable for downstream 3D reasoning. We adopt a pretrained DINOv3-ViTB~\cite{simeoni2025dinov3} backbone, followed by an FPN~\cite{lin2017feature} to build a multi-scale pyramid from surround-view images. Concretely, for an input of 256$\times$704, the backbone generates three multi-scale feature maps. The FPN fuses these features and outputs four feature levels to the BEV lifting stage. The BEV module expects a shared embedding channel (we set $C_{bev} = 256$) and four feature levels, consistent with our encoder configuration.

\textbf{Scene Encoding.} We define the BEV space as a uniform 3D grid spanning $[-51.2, 51.2] \times [-51.2, 51.2] \times [-5.0, 3.0]$ meters in the real world. The scene grid dimensions (default $200 \times 200 \times 16$) determine the number of cells along each axis and allow us to convert grid indices directly into metric coordinates. These coordinates are fed into the BEV positional encoder, where each cell center is encoded using sinusoidal Fourier features and then linearly projected to the BEV feature dimension. This yields the learnable positional embeddings added to the BEV queries before entering the transformer.

The scene encoder receives (i) the BEV queries with positional embeddings and (ii) multi-view image features from four pyramid levels. Before fusion, we attach learned camera embeddings and pyramid-level embeddings to each feature map. The image features are then flattened across views and pyramid levels, and the associated spatial shapes and level start indices are provided to the deformable attention module, which produces the final scene tokens.
\vspace{-3mm}
\subsection{Spatial-Aware Multi-View Decoder}

To mitigate the computational overhead of the multi-view decoder when dealing with long sequences, we first patchify the input BEV features of size $128 \times 128 \times 256$ following the ViT paradigm~\cite{dosovitskiy2020image}. This process yields scene tokens with a reduced spatial resolution of $32 \times 32$ and an embedding dimension of $768$. Subsequently, our spatial-aware multi-view decoder is a ViT-Base~\cite{dosovitskiy2020image} run on a concatenated token sequence: $[\text{CLS}] || \text{BEV}(32 \times 32)||\text{VIEW}(6 \times 16\times 44)$.

\textbf{View Tokens.} View tokens are constructed by starting with a learned mask token for each patch, to which a 2D view positional grid and a learned per-camera embedding are added. For each patch center, Plücker rays are computed using the camera intrinsics and extrinsics, processed with a small MLP, and fused into the view token through a linear projection. 

\textbf{Scene and View Attention.} $Scene \leftrightarrow Scene$ (BEV self-attention) is always enabled to promote global spatial context in BEV, while $View\leftrightarrow View$ (image self-attention) is likewise always enabled across all cameras and patches to encourage cross-view appearance priors.

\begin{figure}[t!]
     \centering
     \includegraphics[width=0.4\linewidth]{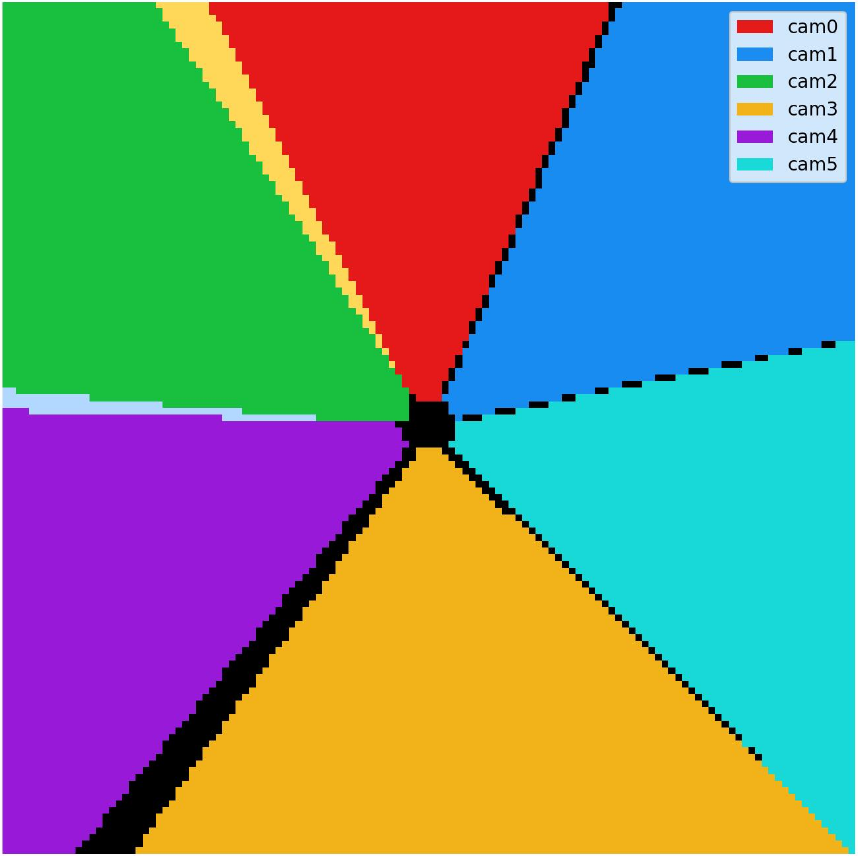}
     \vspace{-1mm}
     \captionof{figure}{\textbf{Visualization of visibility-guided mask.} Each color corresponds to the region in the BEV plane that is visible from a particular camera, forming six wedge-shaped sectors that meet at the ego-vehicle center. The black region at the center corresponds to areas not visible to any camera. In addition, any colors outside these six primary sectors represent boundary zones where multiple camera frustums overlap. }
     \vspace{-3mm}
 \label{fig:vis_mask}
 \end{figure}%

 \begin{figure}[t!]
    \centering
    \includegraphics[width=\linewidth]{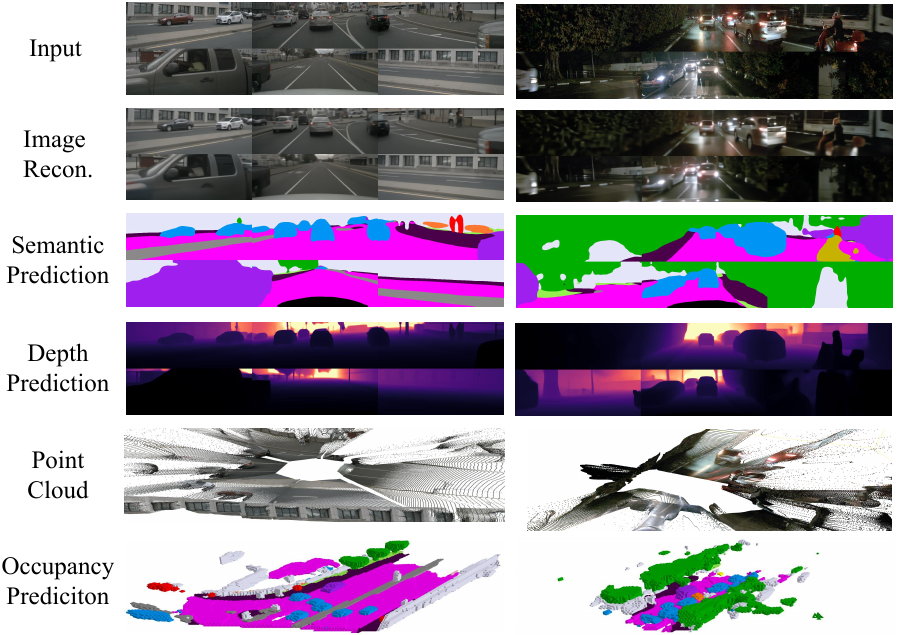}
    \caption{\textbf{Visualizations of DriveTok in different tasks.} Each column corresponds to a different autonomous driving scenario, including challenging night-time scenes. These results illustrate that DriveTok can generate coherent 2D and 3D predictions across diverse tasks from our unified scene tokens.}
    \label{fig:vis3}
    \vspace{-5mm}
\end{figure}

\textbf{Visibility-Guided\;Scene$\leftrightarrow$View\;Attention.} $Scene \leftrightarrow View$ is selectively enabled through a visibility-guided mask: we first compute a boolean tensor $M \in \{0,1\}^{B\times{N_c}\times{N_b}}$ by projecting BEV cell centers into each camera, marking a cell as visible if any height slice lies in front of the camera and falls within the image. This mask is then used to inject an additive bias into the $Scene \leftrightarrow View$ attention logits (in both directions), where entries with $M=0$ indicate that the attention weight between the scene token and the corresponding view should be zero, and the entire mechanism provides hard, geometry-consistent sparsity. We also provide a visualization of our visibility-guided mask in Figure~\ref{fig:vis_mask}. The visualization verifies that the resulting visibility mask captures single-view coverage and non-visible regions, thereby enforcing geometry-consistent sparsity for visibility-guided  $Scene \leftrightarrow View$ attention.
\vspace{-3mm}
\subsection{Joint Task Heads}
\textbf{RGB Reconstruction Head.} The ViewTokenDecoder adopts a DPT~\cite{ranftl2021vision} architecture. Tokens are reshaped into the patch grid and projected to 256 channels with lightweight deblocking. A single-scale feature pyramid is constructed through strided convolutions, yielding four levels $\{256, 512, 1024, 1024\}$ that are refined in a top-down manner using DPT FeatureFusionBlocks. The fused representation is progressively upsampled with additional deblocking to recover full resolution, and a final convolutional head produces the 3-channel RGB output.

\textbf{Depth Prediction Head.} The DepthAdaptor applies a residual MLP to the view tokens and predicts per-camera affine parameters (a, b) to correct monocular scale and shift, after which a second ViewTokenDecoder produces a single-channel depth map using a Softplus activation. The final aligned depth is computed as $a\cdot\text{pred} + b$.

\textbf{Semantics Prediction Head.} A third ViewTokenDecoder with the same architecture is employed for the auxiliary prediction task, producing 17 output channels. Its output remains as raw logits, and the ignore label (255) is excluded during training.

\textbf{Occupancy Prediction Head.} The OccAdaptor maps BEV tokens into a 2D feature plane using LayerNorm and a linear projection, followed by lightweight depthwise--pointwise convolutions to produce a 256-channel feature map at $128^2$. The OccHead upsamples this map with a stride-aligned transposed-convolution stack, while a height-conditioning module provides FiLM parameters across eight vertical slices to lift the 2D features into a 3D volume. Optional 3D refinement blocks further process the volume, and a final $1\times1\times1$ classifier predicts 18 classes. The resulting dense volume is then resampled to the target $200\times200\times16$ occupancy grid.
\vspace{-3mm}
\subsection{Inference Time and Memory Usage of Modules}
To evaluate the efficiency of DriveTok, we measured the average runtime and memory consumption of each module during forward inference on a single NVIDIA A800 80GB GPU, as shown in Table~\ref{tab:efficiency}. In particular, our scene tokenization module is highly efficient ($\sim$ 80-90 ms per scene), enabling real-time processing of surround-view images and producing compact scene tokens that are rich in texture, geometric, and semantic information. Such a lightweight yet expressive representation makes it especially promising for future integration with large language models (LLMs) and vision–language models (VLMs) in autonomous driving systems to support low-latency reasoning.

\vspace{-3mm}
\section{More Visualization Results}
We now provide more visualization results, including challenging night-time scenarios as shown in Figure \ref{fig:vis3}, to better demonstrate the robustness of our approach. We also provide a demo video in the supplementary material. 

\begin{table}[t]
    \centering
    \small
    \caption{\textbf{Average runtime and GPU memory usage of each DriveTok module during forward inference.}}
    \label{tab:drivetok_efficiency}
    \resizebox{0.8\linewidth}{!}{
    \begin{tabular}{lcc}
        \toprule
        \textbf{Module} & \textbf{Avg. time (ms)} & \textbf{Peak GPU memory (MB)} \\
        \midrule
        Image Backbone        & 13.56  & 3747.24  \\
        FPN & 2.01  & 3771.46   \\
        BEV Encoder        & 72.39  & 4423.73  \\
        Spatial-Aware Decoder & 225.85   & 6476.74  \\
        RGB Head &  68.70   & 8456.09 \\
        Depth Head &  69.72   & 8562.64  \\
        Sem Head & 69.18 & 8652.19 \\
        Occ Head &  150.35   & 7802.53  \\
        \midrule
        \textbf{Scene Tokenization} & 87.96 & 4423.73 \\
        \textbf{Total} &  671.76   & 8652.19 \\
        \bottomrule
    \end{tabular}
    }
    \label{tab:efficiency}
\end{table}
\vspace{-3mm}
\section{Dense Depth Annotation}
Public driving datasets such as nuScenes~\cite{caesar2020nuscenes} provide only sparse LiDAR depth, which is insufficient for supervising dense, per-pixel geometry. We obtain dense metric depth by combining monocular predictions with sparse LiDAR supervision. Specifically, MoGe-2~\cite{wang2025moge2} produces a dense pseudo depth map $\tilde{d}$. Sparse LiDAR points are projected onto the image plane to provide metric samples $d^{\text{lidar}}$. We then align $\tilde{d}$ to metric scale using a robust affine fit (ROE):

\begin{equation}
(a^\star, b^\star) = \arg\min_{a,b} \sum_{i\in\Omega} \rho\!\left(a\tilde{d}_i + b - d^{\text{lidar}}_i\right),
\end{equation}
yielding the final aligned dense depth
\begin{equation}
d^{\text{align}}_i = a^\star \tilde{d}_i + b^\star.
\end{equation}
This aligned depth serves as the supervision target for training.

\vspace{-3mm}
\section{More Ablation Studies}
\subsection{Impact of Backbone}
To validate that the performance gains stem from our proposed architectural design rather than solely the powerful backbone, we conduct an ablation study by replacing the DINOv3~\cite{simeoni2025dinov3} encoder with ResNet-101~\cite{he2016deep}, aligning with the configuration of baseline methods. As presented in Table~\ref{tab:exp2}, we strictly limit the input resolution of DriveTok to $256 \times 704$, which is significantly lower than the $896 \times 1600$ resolution used by competing methods. Despite the reduced input information and a standard backbone, our DriveTok (ResNet-101) still demonstrates competitive performance.

\vspace{-4mm}
\begin{table}[t!]
  \centering
  \small
  \setlength{\tabcolsep}{2pt}
\caption{\textbf{Ablation study on backbone configuration.} We compare our DriveTok (equipped with ResNet-101) against state-of-the-art methods using the same backbone. Note that DriveTok uses a significantly lower input resolution ($256 \times 704$) compared to the baselines ($896 \times 1600$) yet achieves competitive results. The red percentages denote the performance drop when switching from DINOv3 to ResNet-101.}

  \label{tab:exp2}
     \vspace{-3mm}
  \resizebox{\linewidth}{!}{
  \begin{tabular}{lcccccccc}
    \toprule
    Method & Backbone & Input & IOU $\uparrow$ & mIOU $\uparrow$ & PSNR $\uparrow$ & SSIM $\uparrow$ & absrel $\downarrow$ & $\delta<1.25$ $\uparrow$\\
    \midrule
    BEVFormer~\cite{li2024bevformer} & ResNet-101 & 896 $\times$ 1600 & 30.50 & 16.75 & - & - & - & - \\
    GaussianFormer~\cite{huang2024gaussianformer} & ResNet-101 & 896 $\times$ 1600 & 29.83  & 19.10  & - & - & - & - \\
    GaussianFormer-2~\cite{huang2025gaussianformer2} & ResNet-101 &  896 $\times$ 1600 & 30.56  & 20.02  & - & - & - & - \\
    QuadricFormer~\cite{zuo2025quadricformer} & ResNet-101 &  896 $\times$ 1600 & 31.22 & \bf{20.12} & - & - & - & - \\
    \midrule
    \textbf{DriveTok} & ResNet-101 & 256 $\times$ 704 & 32.23 (\textcolor{Red}{$\downarrow~$3.2\%}) & 18.83(\textcolor{Red}{$\downarrow~$6.1\%}) & 25.94(\textcolor{Red}{$\downarrow~$6.9\%}) & 0.698(\textcolor{Red}{$\downarrow~$6.6\%}) & 0.10(\textcolor{Red}{$\downarrow~$25.0\%}) & 0.88(\textcolor{Red}{$\downarrow~$5.4\%}) \\
    \textbf{DriveTok} & DINOv3 &  256 $\times$ 704 & \bf33.32 & 20.06 & \bf27.89 & \bf0.747 & \bf0.08 & \bf0.93 \\
    \bottomrule
  \end{tabular} 
}
\vspace{-4mm}
\end{table}

\subsection{Efficiency Analysis}
We compare the inference latency and memory consumption of DriveTok against state-of-the-art occupancy prediction methods in Table~\ref{tab:exp6}. All models use the same ResNet-101 backbone for a fair comparison. While maintaining high-speed inference, our model also achieves competitive performance.

\begin{table}[t!]
  \centering
  \small
  \setlength{\tabcolsep}{8pt}
\caption{\textbf{Latency and memory comparison with Occ models.}}
  \label{tab:exp6}
     \vspace{-3mm}
  \resizebox{\linewidth}{!}{
  \begin{tabular}{lcccccc}
    \toprule
    Method & Backbone & IOU $\downarrow$ & mIOU $\downarrow$ & Latency $\downarrow$ & Peak Memory $\downarrow$ \\
    \midrule
    $\text{GaussianFormer}$~\cite{huang2024gaussianformer} & ResNet-101 & 29.83 & 19.10 & 372 ms &  6229 MB  \\
    $\text{GaussianFormer-2}$~\cite{huang2025gaussianformer2}& ResNet-101 & 30.56 & \bf20.02 & 451 ms & \bf{4535 MB}\\
    \textbf{DriveTok} & ResNet-101 & \bf32.23 & 18.83 &\bf{238 ms} & 7803 MB\\
    \bottomrule
  \end{tabular}
  }
\vspace{-7mm}
\end{table}